\documentclass{article}
\usepackage{spconf,amsmath,graphicx}
\usepackage{multirow}
\usepackage{comment}
\usepackage{booktabs}
\usepackage[table,xcdraw]{xcolor}
\usepackage{colortbl}  
\definecolor{Gray}{gray}{0.85}  

\usepackage{enumitem}
\setlist{nosep, leftmargin=14pt}

\usepackage{mwe} 

\title{Is in-domain data beneficial in transfer learning for landmarks detection in x-ray images?}
\name{Roberto Di Via\textsuperscript{1}, Matteo Santacesaria\textsuperscript{2}, Francesca Odone\textsuperscript{1}, and Vito Paolo Pastore\textsuperscript{1*}\thanks{*Correspondence to Vito.Paolo.Pastore@unige.it}}
\address{\textsuperscript{1} MaLGa Center, DIBRIS,
Università degli studi di Genova, Genoa, Italy, \\ \textsuperscript{2} MaLGa Center, DIMA,
Università degli studi di Genova, Genoa, Italy}
\begin{document}
%
\maketitle


\begin{abstract}
In recent years, deep learning has emerged as a promising technique for medical image analysis.
However, this application domain is likely to suffer from a limited availability of large public datasets and annotations. A common solution to these challenges in deep learning is the usage of a transfer learning framework, typically with a fine-tuning protocol, where a large-scale source dataset is used to pre-train a model, further fine-tuned on the target dataset. 
In this paper, we present a systematic study analyzing whether the usage of small-scale in-domain x-ray image datasets may provide any improvement for landmark detection over models pre-trained on large natural image datasets only. We focus on the multi-landmark localization task for three datasets, including chest, head, and hand x-ray images. 
Our results show that using in-domain source datasets brings marginal or no benefit with respect to an ImageNet out-of-domain pre-training. Our findings can provide an indication for the development of robust landmark detection systems in medical images when no large annotated dataset is available.
\end{abstract}

\begin{keywords}
Landmark detection, Deep learning, Transfer learning, x-ray imaging
\end{keywords}


\begin{figure*}[!ht]
\begin{minipage}[a]{.33\linewidth}
\centering
\includegraphics[width=2.5cm]{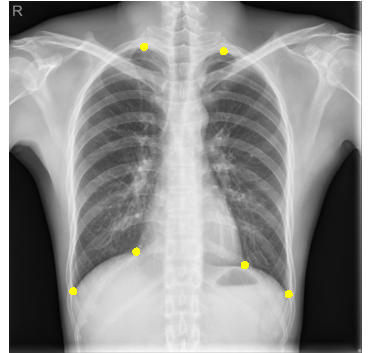}
\centerline{(a) Chest landmarks}\medskip
\end{minipage}
\begin{minipage}[a]{0.33\linewidth}
\centering
\includegraphics[width=2.5cm]{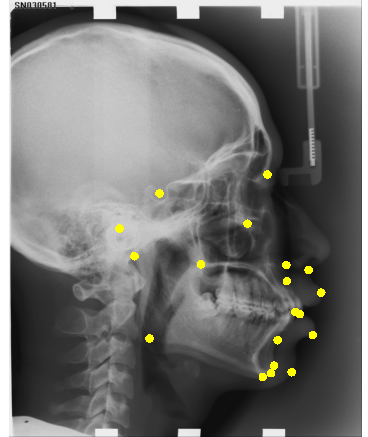}
\centerline{(b) Head landmarks}\medskip
\end{minipage}
\begin{minipage}[a]{0.33\linewidth}
\centering
\includegraphics[width=2.5cm]{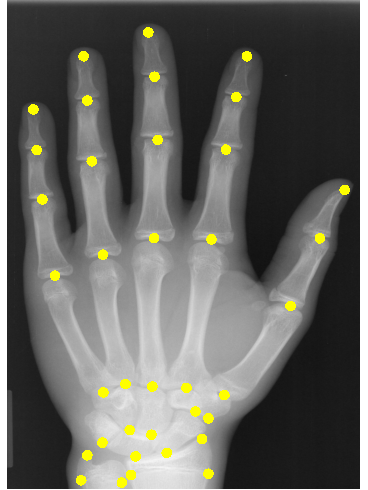}
\centerline{(c) Hand landmarks}\medskip
\end{minipage}
\vspace{-5mm}
\caption{Annotated landmarks in x-ray images for Chest, Head, and Hand datasets.}
\label{fig:data}
\end{figure*}


\section{Introduction}
Landmark localization in medical images is a critical first step in many clinical workflows, enabling key tasks like surgical planning \cite{edwards2021deepnavnet}. Manually identifying landmarks on medical images is a tedious and time-consuming process, subject to high inter-operator variability. Hence, there is a need for robust automated landmark detection systems to improve clinical efficiency and standardization \cite{DBLP:journals/tmi/WuLLPPLWWCWJCSK22}. As a consequence, deep learning has been extensively applied to the task at hand. X-ray imaging is one of the most common and economical modalities used for anatomical assessment and diagnosis. 
Landmark detection in x-rays can directly support skeletal measurement \cite{DBLP:journals/ijon/LeeCS22}. Several works have focused on designing and applying deep learning models to landmark localization in specific anatomical regions like the chest, head, or hand, from x-ray images \cite{zhu2021you}. However, deep learning is notoriously data-hungry, and the limited availability of training samples may be problematic for its application to the medical domain, including anatomical landmark detection in x-ray images. A popular and widely adopted solution to this issue is the usage of a transfer learning framework \cite{DBLP:conf/sac/BaeAHKPC21}. 
While it is known from the literature that the number of images and classes of the dataset used as a source in a transfer learning framework directly influences the performance of the resulting model on the target dataset \cite{alfano2024top}, there is still limited literature comparing the impact of in-domain over natural image pre-training datasets \cite{zamir2018taskonomy,touijer2023food}, with contrasting results depending on the specific domain and downstream task.  
In an attempt to fill this gap in the context of landmark detection from x-ray images, in this paper, we present a systematic study on the effect of in-domain transfer learning, analyzing whether small-scale in-domain datasets may provide any improvement for landmark detection over models pre-trained on large natural image datasets only. We develop a deep learning pipeline based on the U-Net++ architecture for multi-landmark localization in chest, head, and hand x-ray images, reaching state-of-the-art results for three benchmark datasets. Comprehensive experiments are conducted to compare the effect of in-domain fine-tuning on models pre-trained on ImageNet, prior to fine-tuning on the target datasets.
To our knowledge, this is the first work that evaluates the usefulness of in-domain training data for anatomical landmark localization across multiple x-ray imaging domains. Our findings can inform the development of generalized landmark detection systems in medical images, suggesting that they may not benefit from small-scale annotated in-domain datasets. 
The remainder of the paper is organized as follows: in Section \ref{Rel_works} we provide a description of related works, in Section \ref{methods} we present the proposed pipeline and experimental framework, while in Section \ref{RESULTS} we provide the experiment details and present the obtained results, drawing some conclusions in Section \ref{conclusions}.

\section{Related Works}\label{Rel_works}

In this Section, we report a description of relevant works in the context of automatic anatomical landmark detection in x-ray images. 
Tiulpin et al. \cite{DBLP:journals/corr/abs-1907-12237} introduce a landmark localization method for knee x-rays in osteoarthritis stages, treating it as a regression problem. Their hourglass architecture captures multi-scale features. They explore regularization techniques and loss functions' impact and employ transfer learning from low-budget annotations, significantly boosting accuracy.
Yeh et al. \cite{yeh2021deep} develop a deep learning model detecting 45 landmarks and 18 parameters on spine x-rays, revealing varying errors across spinal regions. The deep learning model is derived from a modified Cascaded Pyramid Network, applying a two-stage, coarse-to-fine approach for landmark localization. 
Zhu et al. \cite{zhu2021you} propose GU2Net, a universal model for anatomical landmark detection in medical images, addressing limitations in specialized, dataset-dependent methods. GU2Net is composed of two main parts: a local network for learning multi-domain local features, and a global network for extracting global features to clarify landmark locations further, achieving superior performance with fewer parameters across head, hand, and chest x-ray datasets containing 62 landmarks.
Differently from the referenced works, in this paper, we tackle the data scarcity problem performing a systematic study on transfer learning for landmark detection in x-ray images. Specifically, we evaluate the effect of small-scale in-domain datasets for the fine-tuning of ImageNet pre-trained neural networks, in terms of performances on three publicly available x-ray image datasets.

\section{Method}
\label{methods}

\subsection{Datasets description}

In this section, we describe the three annotated public x-ray datasets used in this study: chest, head, and hand. Annotated sample images for each dataset are shown in Fig. \ref{fig:data}. For benchmarking our results, we consider the same settings of \cite{zhu2021you}, including image selection, training, and test data criteria. \\

\noindent
\textbf{CHEST.} The Chest dataset includes 279 chest x-ray images from the China set \cite{DBLP:journals/tmi/CandemirJPMSXKATM14} collected from the U.S. National Library of Medicine's public repository. The images are approximately 3000x3000 pixels. 
We use the 6 landmarks manually annotated in each image provided in \cite{zhu2021you}, exploiting the first 229 images for training and the remaining 50 for testing. Pixel distance is used as the evaluation metric.

\noindent
\textbf{HEAD.} The Head dataset \cite{DBLP:journals/mia/WangHLLCSLIVRFC16} is an open-source collection comprising 400 cephalometric x-ray images representing individuals spanning ages from 7 to 76 years. These images have a resolution of 0.1mm x 0.1mm with 2400x1935 pixels. In our study, we employ the initial 150 images for training and the remaining 250 images for testing (as done in \cite{zhu2021you}). To ensure precise anatomical landmark annotations, two orthodontists marked 19 landmarks on each image. Mean coordinates were computed to account for inter-observer variations.

\noindent
\textbf{HAND.} The Hand dataset is a publicly available collection of 909 hand x-ray images, with an average size of 1563x2169 pixels. In our study, we use the first 609 images for training and the remaining 300 images for testing (as done in \cite{zhu2021you}). To calculate the physical distance and compare it with other methods, we assume that the width of the wrist is 50mm, following Payer et al.'s proposal \cite{DBLP:journals/mia/PayerSBU19}. The physical distance is computed by multiplying the pixel distance with $\frac{50}{\|p-q\|_2}$, where $p$ and $q$ are the two endpoints of the wrist. Payer et al. \cite{DBLP:journals/mia/PayerSBU19} have manually labeled a total of 37 landmarks, and the two endpoints of the wrist can be obtained from the first and fifth points, respectively. \\
\vspace{-4mm}
\begin{figure}[!ht]
\centering
\includegraphics[width=1.0\columnwidth]{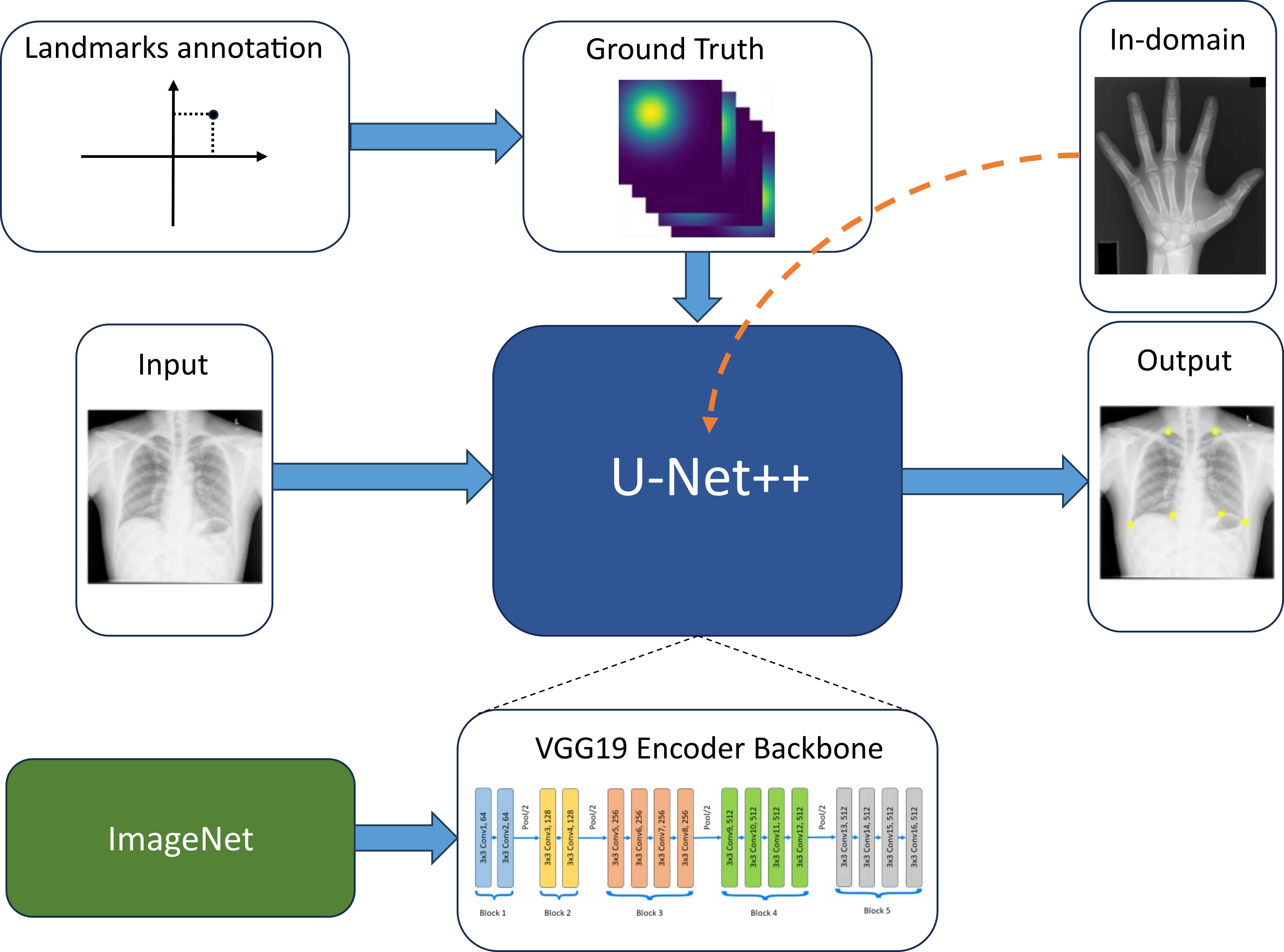}
\caption{Schematic representation of our pipeline. The orange arrow indicates the additional in-domain fine-tuning.}
\label{fig:pipeline}
\end{figure}
\vspace{-5mm}

\subsection{Proposed Pipeline}

In this Section, we introduce the proposed pipeline, providing details about each of its components. A schematic representation of the pipeline is shown in Fig. \ref{fig:pipeline}. First, expert-annotated landmarks are used to generate Gaussian heatmaps representing ground truth labels, where each landmark is represented by a single heatmap. Next, the medical images and the generated ground truth labels are augmented and fed into the U-Net++ model \cite{DBLP:journals/corr/abs-1807-10165}, which uses an ImageNet pre-trained VGG19 encoder to extract image features. The U-Net++ architecture is a popular choice for medical image segmentation tasks, as it is able to learn spatial relationships between pixels and produce accurate segmentation results. The output of the trained model is heatmaps predicting landmark locations, further combined into a mask localizing all landmarks. \\

\vspace{-5mm}
\section{Experiments}\label{RESULTS}

\subsection{Experiment details}
\label{sec:experiments_details}
We use the Pytorch library for the development of the proposed pipeline. To determine the optimal architecture and backbone while evaluating the robustness of the resulting model, we conduct a 5-fold cross-validation procedure.
All models are trained with Adam optimizer for 200 epochs, with an early stopping criterion. The learning rate is adjusted using a ReduceLROnPlateau scheduler based on the validation loss, with an initial value of 0.0001 and the batch size is fixed to 2. In our protocol, all the layers are fine-tuned. The images and annotations of the chest, head, and hand x-ray datasets are augmented and resized to 512x512 for training and 512x512, 512x416, and 512x368 respectively for testing, as exhibited in the experiments of \cite{zhu2021you}. 

\subsection{Evaluation Metrics}

We adopt the same evaluation metrics of \cite{zhu2021you}, that are the Mean Radial Error (MRE) measured in pixels or mm, and Success Detection Rate (SDR) at different threshold distances. \\
The MRE is the average Euclidean distance between the predicted landmarks and the ground truth landmarks. In the case of $n$ landmarks, the MRE is calculated as:
$$MRE = \frac{1}{n} \sum_{i=1}^{n} \sqrt{(x_i - \hat{x}_i)^2 + (y_i - \hat{y}_i)^2}$$
where $(x_i, y_i)$ are the coordinates of the ground truth landmark and $(\hat{x}_i, \hat{y}_i)$ are the coordinates of the predicted ones. \\
The SDR measures the percentage of predicted landmarks that are within a threshold distance of the ground truth. In the case of $n$ landmarks and a threshold $t$, the SDR is calculated as:
$$SDR = \frac{1}{n} \sum_{i=1}^{n} I(d_i \leq t)$$
where $d_i$ is the Euclidean distance between the ground truth and the predicted landmark, $I$ is the indicator function that returns 1 if $d_i <= t$ and 0 otherwise.

\vspace{-2mm}
\subsection{Comparative Study on Architecture and Backbones}

In this Section, we assess a variety of models on the hand dataset (as it has the highest number of landmarks), to identify the best-performing model for the transfer learning experiment, while establishing a baseline. The investigated architectures include U-Net \cite{DBLP:journals/corr/RonnebergerFB15}, U-Net++ \cite{DBLP:journals/corr/abs-1807-10165}, and DeepLabV3 \cite{DBLP:journals/corr/ChenPSA17}. The results are summarized in Table \ref{table:model_selection}, reporting the average metrics computed with a 5-fold cross-validation procedure (see Sec. \ref{sec:experiments_details}). For all the architectures, we consider three different ImageNet pre-trained encoder backbones: EfficientNet-B7, ResNeXt101, and VGG19, with the exception of DeepLabV3, which does not support the VGG19 encoder. U-Net++ with a VGG19 encoder pre-trained on ImageNet achieves the best MRE of 0.50mm and SDR of 98.90\% at the most restrictive threshold (2mm). VGG19 outperforms the other encoders, reducing MRE by 50\% with U-Net and 37.5\% with U-Net++ compared to EfficientNet-B7. Among DeepLabV3 models, ResNeXt101 obtains the lowest MRE of 1.00mm but is not as successful as U-Net++/VGG19. According to these results, in the next experiments, we use a Unet++ architecture with a VGG19 backbone. 
\vspace{-3mm}
\begin{table}[!ht]
  \centering
  \caption{Performance comparison of models and encoder backbones for x-ray anatomical landmark detection. Results are from a 5-fold cross-validation approach. The best results are highlighted in bold, the second-best results are underlined.}
  \label{table:model_selection}
  \resizebox{1.0\columnwidth}{!}{%
  \begin{tabular}{ l l c c c c c }
    \hline
     \multirow{3}{*}{Models} & \multirow{3}{*}{Backbones} & \multicolumn{4}{c}{\textbf{Hand}} \\
              & & MRE±STD $\downarrow$ & \multicolumn{3}{c}{SDR(\%) $\uparrow$} \\
    \cline{4-6}
    & & (mm) & 2mm & 4mm & 10mm \\
    
    
    \hline
    \multirow{3}{*}{U-net \cite{DBLP:journals/corr/RonnebergerFB15}}
                    & efficientnet-b7    & 1.20 $\pm$ 0.52 & 95.44 $\pm$ 1.74 & 99.00 $\pm$ 0.82 & 99.28 $\pm$ 0.71 \\
                    & resnext101\_32x8d  & 0.79 $\pm$ 0.11 & 96.64 $\pm$ 0.40 & 99.60 $\pm$ 0.23 & 99.78 $\pm$ 0.16 \\
                    & vgg19              & \underline{0.60 $\pm$ 0.02} & \underline{97.38 $\pm$ 0.34} & \underline{99.82 $\pm$ 0.10} & \underline{99.98 $\pm$ 0.02} \\
    \hline
    \multirow{3}{*}{U-net++ \cite{DBLP:journals/corr/abs-1807-10165}} 
                    & efficientnet-b7   & 0.80 $\pm$ 0.12 & 96.60 $\pm$ 0.77 & 99.52 $\pm$ 0.24 & 99.76 $\pm$ 0.17 \\ 
                    & resnext101\_32x8d & 0.74 $\pm$ 0.05 & 96.21 $\pm$ 0.31 & 99.61 $\pm$ 0.08 & 99.86 $\pm$ 0.11 \\
                    & vgg19             & \textbf{0.50 $\pm$ 0.12} & \textbf{98.90 $\pm$ 0.85} & \textbf{99.88 $\pm$ 0.17} & 99.94 $\pm$ 0.08 \\
    \hline
    \multirow{3}{*}{DeepLabV3 \cite{DBLP:journals/corr/ChenPSA17}} 
                   & efficientnet-b7  & 1.35 $\pm$ 0.01 & 80.70 $\pm$ 0.20 & 98.24 $\pm$ 0.29 & 99.97 $\pm$ 0.01 \\ 
                   & resnext101\_32x8d & 1.00 $\pm$ 0.05 & 89.93 $\pm$ 1.09 & 99.35 $\pm$ 0.23 & \textbf{99.99 $\pm$ 0.01} \\ 
                   & vgg19            & -               & -                & -                & -                \\ 
    \hline
  \end{tabular}%
  }
\end{table}

\begin{table*}[!ht]
  \centering
  \caption{Comparison of transfer learning strategies for anatomical landmarks detection in x-ray images. The best results are
highlighted in bold, and the second-best results are underlined. }
  \label{table:transfer_learning}
  \resizebox{0.8\textwidth}{!}{%
   \begin{tabular}{l | c c c c | c c c c c | c c c c}
      \hline
     \textbf{Target dataset} & \multicolumn{4}{c|}{$\rightarrow$\textbf{Chest}} & \multicolumn{5}{c|}{$\rightarrow$\textbf{Head}} & \multicolumn{4}{c}{$\rightarrow$\textbf{Hand}} \\
     \hline 
       \multirow{2}{*}{\textbf{Model weights}} & MRE $\downarrow$ & \multicolumn{3}{c|}{SDR(\%) $\uparrow$} & MRE $\downarrow$ & \multicolumn{4}{c|}{SDR(\%) $\uparrow$} & MRE $\downarrow$ & \multicolumn{3}{c}{SDR(\%) $\uparrow$} \\
       & (px) & 3px & 6px & 9px & (mm) & 2mm & 2.5mm & 3mm & 4mm & (mm) & 2mm & 4mm & 10mm \\
    \hline
  
    Imagenet & \textbf{4.19} & \textbf{46.75} & \textbf{77.24} & \textbf{89.84} & 1.50 & 77.79 & 85.33 & 91.01 & 96.48 & \textbf{0.65} & \textbf{96.90} & \textbf{99.84} & \underline{99.95} \\ 

    Imagenet $\rightarrow$ Chest & - & - & - & - & \textbf{1.45} & \textbf{79.39} & \textbf{86.13} & \textbf{92.53} & \textbf{97.28} & 0.81 & \underline{96.72} & 99.60 & 99.74 \\ 
    Imagenet $\rightarrow$ Head  & 4.78 & 38.21 & 71.14 & 87.40 & - & - & - & - & - & 0.84 & 95.77 & 99.49 & 99.75 \\ 
    Imagenet $\rightarrow$ Hand  & 5.09 & 35.77 & 69.11 & 86.99 & \underline{1.49} & 78.27 & \underline{85.79} & \underline{91.43} & 96.48 & - & - & - & - \\ 

    Imagenet $\rightarrow$ Chest $\rightarrow$ Head & - & - & - & - & - & - & - & - & - & 0.76 & 95.95 & 99.71 & 99.89 \\ 
    Imagenet $\rightarrow$ Chest $\rightarrow$ Hand & - & - & - & - & 1.57 & 75.18 & 83.01 & 89.94 & 95.94 & - & - & - & - \\ 

    Imagenet $\rightarrow$ Head $\rightarrow$ Chest & - & - & - & - & - & - & - & - & - & \underline{0.67} & 96.35 & \underline{99.82} & \textbf{99.97} \\ 
    Imagenet $\rightarrow$ Head $\rightarrow$ Hand  & 4.89 & 38.21 & 71.54 & \underline{88.21} & - & - & - & - & - & - & - & - & - \\ 

    Imagenet $\rightarrow$ Hand $\rightarrow$ Chest & - & - & - & - & 1.50 & \underline{78.44} & 85.64 & 91.07 & \underline{96.67} & - & - & - & - \\ 
    Imagenet $\rightarrow$ Hand $\rightarrow$ Head  & \underline{4.86} & \underline{43.90} & \underline{71.95} & 84.96 & - & - & - & - & - & - & - & - & - \\ 
    \hline
    \end{tabular}
    }
    \vspace{-5mm}
\end{table*}

\begin{table*}[!ht]
  \centering
  \caption{Comparison of state-of-the-art results for anatomical landmark detection in x-ray images.}
  \label{table:state-of-the-art}
  \resizebox{0.75\textwidth}{!}{%
  \begin{tabular}{ l | c c c c | c c c c c | c c c c }
    \hline
         \multirow{3}{*}{Methods} & \multicolumn{4}{c|}{\textbf{Chest}} & \multicolumn{5}{c|}{\textbf{Head}} & \multicolumn{4}{c}{\textbf{Hand}} \\
     & MRE $\downarrow$ & \multicolumn{3}{c|}{SDR(\%) $\uparrow$} & MRE $\downarrow$ & \multicolumn{4}{c|}{SDR(\%) $\uparrow$} & MRE $\downarrow$ & \multicolumn{3}{c}{SDR(\%) $\uparrow$}  \\
    \cline{3-5} \cline{7-10} \cline{12-14}
    & (px) & 3px & 6px & 9px & (mm) & 2mm & 2.5mm & 3mm & 4mm & (mm) & 2mm & 4mm & 10mm \\
    \hline
    Lindner et al. \cite{Lindner2016}                & -    & -     & -     & -     & 1.67 & 70.65 & 76.93 & 82.17 & 89.85 & 0.85 & 93.68 & 98.95 & 99.94 \\
    Urschler et al. \cite{articleUrschler}           & -    & -     & -     & -     & -    & 70.21 & 76.95 & 82.08 & 89.01 & 0.80 & 92.19 & 98.46 & 99.95 \\
    Payer et al. \cite{DBLP:journals/mia/PayerSBU19} & -    & -     & -     & -     & -    & 73.33 & 78.76 & 83.24 & 89.75 & \underline{0.66} & 94.99 & 99.27 & \textbf{99.99} \\
    Zhu et al. \cite{zhu2021you}                     & 5.57 & \textbf{57.33} & \textbf{82.67} & \underline{89.33} & \underline{1.54} & \textbf{77.79} & \underline{84.65} & \underline{89.41} & \underline{94.93} & 0.84 & \underline{95.40} & \underline{99.35} & 99.75 \\
    Our Proposed Pipeline                            & \textbf{4.19} & \underline{46.75} & \underline{77.24} & \textbf{89.84} & \textbf{1.50} & \textbf{77.79} & \textbf{85.33} & \textbf{91.01} & \textbf{96.48} & \textbf{0.65} & \textbf{96.90} & \textbf{99.84} & \underline{99.95} \\
    \hline

  \end{tabular}%
  }
\end{table*}
\vspace{-5mm}
\subsection{The effect of in-domain fine-tuning on x-rays landmark detection}
At this stage, we evaluate whether the usage of small-scale in-domain dataset transfer learning brings any further benefit in terms of performance on the test set, with respect to our best model with an ImageNet pre-trained backbone (Unet++/VGG19). 
In an attempt to provide insights in this direction, we evaluate multiple stages of in-domain fine-tuning, considering all the possible transfer combinations between the three datasets investigated in this work. For instance, if we consider the chest dataset as a target, we can fine-tune the ImageNet pre-trained model on the head dataset or on the hand dataset, prior to actually fine-tuning such model on the chest dataset. We dig further into the benefit of in-domain transfer learning, allowing two stages of fine-tuning. Intuitively, considering the chest dataset as a target, we perform the following steps: (i) fine-tuning the ImageNet pre-trained model on the hand dataset; (ii) fine-tuning the resulting model on the head dataset; (iii) fine-tuning the resulting model on the chest dataset. The steps (i) and (ii) can be inverted to evaluate the effect of such permutation on the results. We repeat the designed two-stage procedure considering each one of the three datasets included in this work, as a target. Table \ref{table:transfer_learning} summarizes the obtained results. In this table, the operator $\rightarrow$ refers to the fine-tuning procedure, so that $ImageNet \rightarrow A$ in the \textit{Model weights} entry, means that the 
Unet++ model with VGG19 ImageNet pre-trained backbone has been fine-tuned on A, with the resulting model further fine-tuned on the target dataset. $ImageNet \rightarrow A \rightarrow B$ refers to the two-stage fine-tuning procedure.  
The baseline obtained by directly fine-tuning the ImageNet pre-trained model on the target dataset is reported in Table \ref{table:transfer_learning} as well, as a reference.\\
\noindent
Our results show that fine-tuning the ImageNet pre-trained models on in-domain small-scale datasets generally does not lead to significant performance gains, but in the case of the head dataset, when fine-tuning the ImageNet pre-trained model on the chest dataset prior to transfer on the head dataset, provides an improvement of $0.05$ in MRE and $1.60\%$ in SDR at the 2mm threshold. In some cases, such as the SDR at 10mm for hand x-rays, in-domain fine-tuning even decreases the performance. 
These results suggest that the feature representations learned from ImageNet transfer effectively to this new domain, and the in-domain small-scale datasets are not enough to provide additional improvements, providing empirical evidence that ImageNet pretraining should be the primary transfer learning approach for landmark detection tasks in x-ray images. 
\vspace{-3mm}
\subsection{Comparison with the state-of-the-art}
\vspace{-2mm}
In this Section, we compare the performances of the Imagenet pipeline against several state-of-the-art methods for landmark detection in chest, head, and hand x-ray datasets. Table \ref{table:state-of-the-art} reports the obtained results. On chest x-rays, our pipeline achieves the lowest MRE of 4.19 pixels, outperforming Zhu et al. \cite{zhu2021you} by 25\%. At the 9px threshold, we attain the best SDR of 89.84\%. For head x-rays, our pipeline obtains the lowest MRE of 1.50mm, beating Lindner et al. \cite{Lindner2016} by 10\%. Our pipeline achieved the highest SDRs with 3-7\% improvement at 2.5-4mm thresholds. On the hand dataset, we obtain the lowest MRE of 0.65mm, slightly better than Payer et al. \cite{DBLP:journals/mia/PayerSBU19}. At clinically important 2mm and 4mm thresholds, our pipeline achieves the highest SDRs of 96.90\% and 99.84\%. 
\vspace{-4mm}
\section{Conclusion}\label{conclusions}
\vspace{-2mm}
In this work, we present a study analyzing the benefits of small-scale in-domain datasets for landmark detection across diverse x-ray datasets. We design a deep learning pipeline based on a U-Net++ with an ImageNet VGG19 pre-trained encoder and evaluate chest, head, and hand x-rays. Our experiments show that in-domain fine-tuning does not substantially improve over ImageNet pre-training. This suggests that rich features from natural images effectively transfer to x-ray images without needing further in-domain data. Our models achieve state-of-the-art landmark localization accuracy, obtaining the lowest MREs and highest SDRs at clinically important thresholds. Our findings show that ImageNet pretraining is the most effective strategy for landmark detection in diverse x-ray datasets, guiding the development of robust systems without needing large-scale annotated in-domain data.




\vspace{-3mm}
\section{Compliance with ethical standards}
\label{sec:ethics}
\vspace{-3mm}
This research study was conducted retrospectively using human subject data made available in open access. Ethical approval was not required as confirmed by the license attached with the open-access data.
\vspace{-3mm}

\bibliographystyle{IEEEbib}

\end{document}